# Semi-parametric Image Synthesis

Xiaojuan Qi  Qifeng Chen  Jiaya Jia  Vladlen Koltun
CUHK         Intel Labs   CUHK      Intel Labs

## Abstract

*We present a semi-parametric approach to photographic image synthesis from semantic layouts. The approach combines the complementary strengths of parametric and nonparametric techniques. The nonparametric component is a memory bank of image segments constructed from a training set of images. Given a novel semantic layout at test time, the memory bank is used to retrieve photographic references that are provided as source material to a deep network. The synthesis is performed by a deep network that draws on the provided photographic material. Experiments on multiple semantic segmentation datasets show that the presented approach yields considerably more realistic images than recent purely parametric techniques.*

## 1. Introduction

*Zeuxis having painted a child carrying grapes, the birds came to peck at them; upon which [...] he expressed himself vexed with his work, and exclaimed – "I have surely painted the grapes better than the child, for if I had fully succeeded in the last, the birds would have been in fear of it."*

– Pliny the Elder, The Natural History, 79 AD

Photographic image synthesis by deep networks can open a new route to photorealism: a problem that has traditionally been approached via explicit manual modeling of three-dimensional surface layout and reflectance distributions [24]. A deep network that is capable of synthesizing photorealistic images given a rough specification could become a new tool in the arsenal of digital artists. It could also prove useful in the creation of AI systems, by endowing them with a form of visual imagination [19].

Recent progress in photographic image synthesis has been driven by parametric models – deep networks that represent all data concerning photographic appearance in their weights [11, 2]. This is in contrast to the practices of human photorealistic painters, who do not draw purely on memory but use external references as source material for reproducing detailed object appearance [17]. It is also in contrast to earlier work on image synthesis, which was based on nonparametric techniques that could draw on large datasets of images at test time [7, 15, 3, 13, 10]. In switching from nonparametric approaches to parametric ones, the research community gained the advantages of end-to-end training of highly expressive models. But it relinquished the ability to draw on large databases of original photographic content at test time: a strength of earlier nonparametric techniques.

In this paper, we present a semi-parametric approach to photographic image synthesis from semantic layouts. The presented approach exemplifies a general family of methods that we call semi-parametric image synthesis (SIMS). Semi-parametric synthesis combines the complementary strengths of parametric and nonparametric techniques. In the presented approach, the nonparametric component is a database of segments drawn from a training set of photographs with corresponding semantic layouts. At test time, given a novel semantic layout, the system retrieves compatible segments from the database. These segments are used as raw material for synthesis. They are composited onto a canvas with the aid of deep networks that align the segments to the input layout and resolve occlusion relationships. The canvas is then processed by a deep network that produces a photographic image as output.

We conduct experiments on the Cityscapes, NYU, and ADE20K datasets. The experimental results indicate that images produced by SIMS are considerably more realistic than the output of purely parametric models for photographic image synthesis from semantic layouts.

## 2. Related Work

Recent work on conditional image synthesis is predominantly based on parametric models [26, 34, 32, 25, 6, 21, 22, 8, 11, 2, 37]. Most related to ours are the works of Isola et al. [11] and Chen and Koltun [2]. Isola et al. propose a general framework for image-to-image translation based on adversarial training. This approach can be applied to synthesize images from semantic layouts. Chen and Koltun propose a direct approach to synthesizing high-resolution images conditioned on semantic layouts. Their method does not rely on adversarial training, but rather trains a convolutional network directly with a perceptual loss. Our approach differs from all of these in that a memory bank of object segments is utilized at test time as source material for synthesis. Synthesis is performed by a deep network, but is based on exemplars of object appearance retrieved from the memory bank. Figure 1 provides a qualitative comparison.

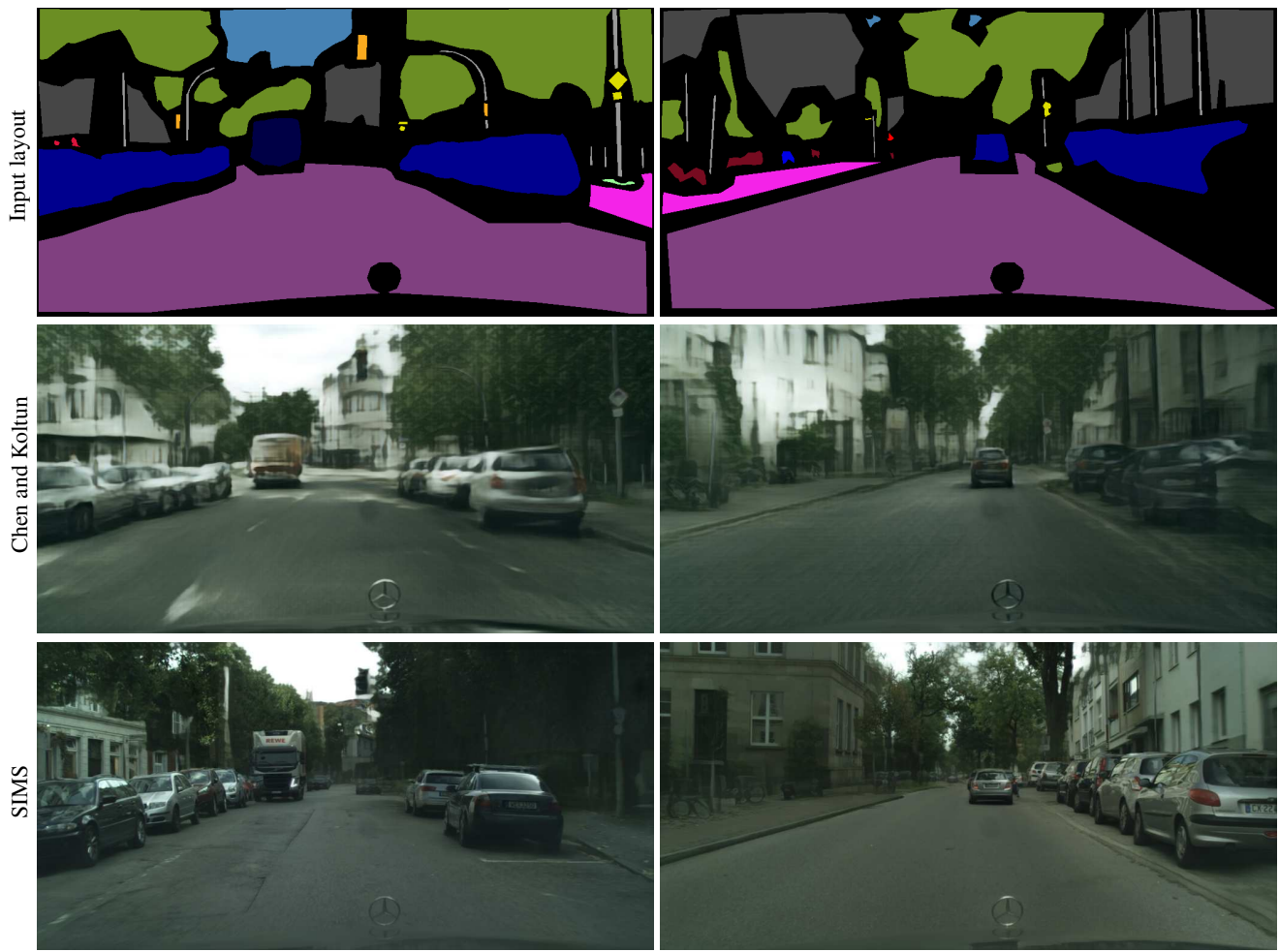

Figure 1. Comparison to the approach of Chen and Koltun [2] on coarse semantic layouts from the Cityscapes dataset. Zoom in for details.

Nonparametric methods for image synthesis have a long history and were dominant before the ascendance of purely parametric techniques. Hays and Efros [7] used a collection of images as source material for image completion. At test time, similar images are retrieved via descriptor matching and are used to inpaint missing regions. Lalonde et al. [15] developed an interactive system that retrieves object segments from a large library of images. The retrieved segments are interactively composited onto an image. Chen et al. [3] described a system that synthesized an image from a freehand sketch with associated text labels. Given a sketch and associated text, their system retrieves relevant images from the Web, segments them, and composes an output image with interactive assistance by the user. Johnson et al. [13] described a related system for post-processing computer-generated images. Isola and Liu [10] presented an analysis-by-synthesis approach that retrieves object segments that match a query image and combines these segments to form a "scene collage" that explains the query. Our research is inspired by this line of work and aims to reintroduce these earlier ideas into the current stream of image synthesis research. Unlike the earlier work, our approach combines nonparametric use of a database of image segments with deep parametric models that assist composition and perform synthesis based on the retrieved material.

Zhu et al. [40] train a convolutional network to predict the realism of image composites. (See also the earlier work of Lalonde and Efros [14] and Xue et al. [33].) Tsai et al. [31] train a convolutional network to harmonize the appearance of image composites. In these works, the composites were assumed to be given, typically generated interactively by a human user. In contrast, our work develops a complete automatic pipeline for semi-parametric image synthesis from semantic layouts.

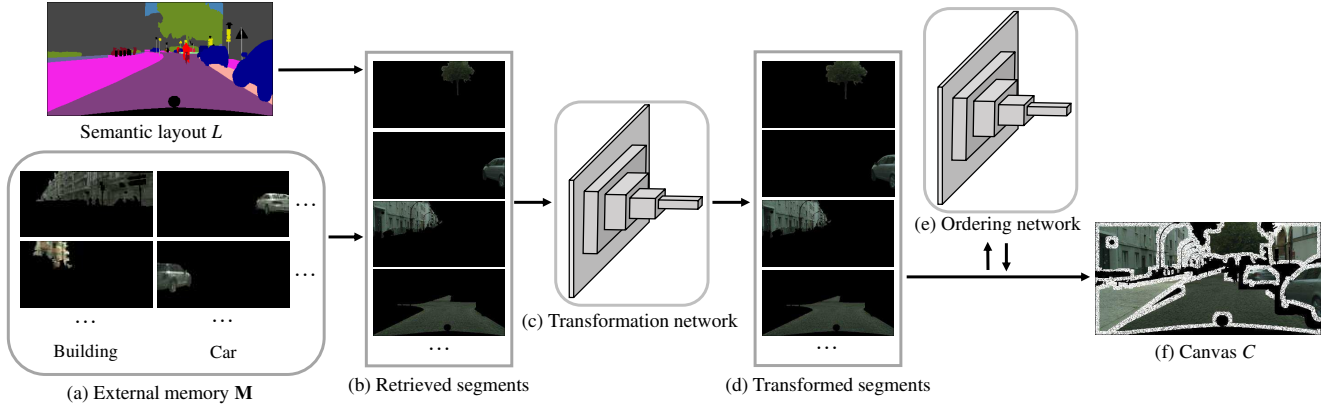

Figure 2. First stage of the image synthesis pipeline. (a) Given a semantic layout $L$, the external memory $\mathbf{M}$ is queried to retrieve compatible segments (b), which are aligned to the input layout by a spatial transformer network (c,d). An ordering network (e) assists the composition of the canvas $C$ (f). The synthesis process continues as illustrated in Figure 3.

## 3. Overview

Our goal is to synthesize a photorealistic image based on a semantic layout $L \in \{0,1\}^{h \times w \times c}$, where $h \times w$ is the image size and $c$ is the number of semantic classes. Our model is trained on a set of paired color images and corresponding semantic layouts. This set is used to generate a memory bank $\mathbf{M}$ of image segments from different semantic categories. Segments are extracted from training images by taking connected components in corresponding semantic layouts. Each segment $P_i$ in $\mathbf{M}$ is a segment from a training color image, associated with a semantic class. A number of segments are shown in Figure 2(a,b).

At test time, we are given a semantic label map $L$ that was not seen during training. This label map is decomposed into connected components $\{L_i\}$. For each connected component, we retrieve a compatible segment from $\mathbf{M}$ based on shape, location, and context (Figure 2(b)). This retrieved segment is aligned to $L_i$ by a spatial transformer network trained for this purpose [12] (Figure 2(c,d)). The transformed segments are composited onto a canvas $C \in \mathbb{R}^{w \times h \times 3}$ (Figure 2(f)). Since the segments may not align perfectly with the masks $\{L_i\}$, they may overlap. Relative front-back order is determined by an ordering network (Figure 2(e)). Boundaries of retrieved segments are deliberately elided. The composition of the canvas $C$ is described in detail in Section 4.

The canvas $C$ and the input layout $L$ are used as input to a synthesis network $f$. This network synthesizes the final output image and is illustrated in Figure 3. It inpaints missing regions, harmonizes retrieved segments, blends boundaries, synthesizes shadows, and otherwise adjusts and synthesizes photographic appearance based on the raw material in the canvas $C$ and the target layout $L$. The architecture and training of the network $f$ are described in Section 5.

To apply the presented approach to coarse input layouts, such as ones shown in Figure 1, we train a cascaded refinement network to convert coarse incomplete layouts to dense pixelwise layouts [2]. The network is trained on pairs of coarse and fine semantic layouts. At test time, given a coarse incomplete layout, the trained network synthesizes a dense semantic layout, which is then provided to the presented image synthesis pipeline.

## 4. External Memory

### 4.1. Representation

The memory bank $\mathbf{M}$ is a set of image segments $\{P_i\}$ extracted from the training data. Each segment corresponds to a maximal connected component in the semantic label map of one of the training images. A segment $P_i$ is associated with a tuple $(P_i^{color}, P_i^{mask}, P_i^{cont})$, where $P_i^{color} \in \mathbb{R}^{h \times w \times 3}$ is a color image that contains the segment (other pixels are zeroed out), $P_i^{mask} \in \{0,1\}^{h \times w \times c}$ is a binary mask that specifies the segment's footprint, and $P_i^{cont} \in \{0,1\}^{h \times w \times c}$ is a semantic map representing the semantic context around $P_i$ within a bounding box, obtained from the semantic label map that originally contained the segment. The bounding box that encloses the context region is obtained by computing the bounding box of $P_i^{color}$ and enlarging it by 25% in each dimension.

### 4.2. Retrieval

Given a novel semantic layout $L$ at test time, we compute $L_j^{mask}$ and $L_j^{cont}$ for each semantic segment $L_j$, by analogy with the definitions provided in Section 4.1. Then for each segment $L_j$ in the test image $L$, we select the most compatible segment $P_{\sigma(j)}$ in $\mathbf{M}$ based on a similarity score:

$$\sigma(j) = \arg\max_i \text{IoU}(P_i^{mask}, L_j^{mask}) + \text{IoU}(P_i^{cont}, L_j^{cont}),$$

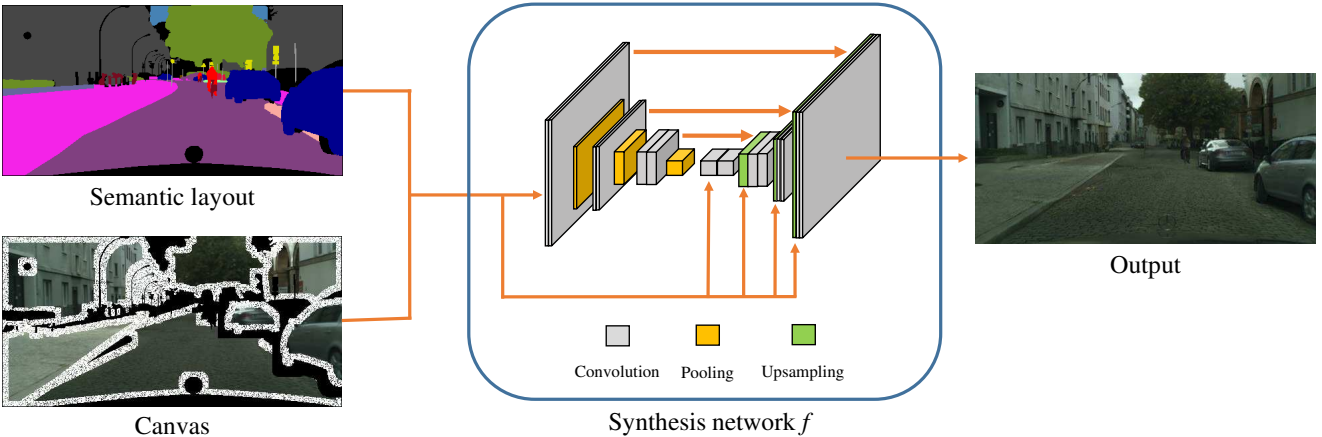

Figure 3. The target semantic layout $L$ and the canvas $C$ are given to a network $f$, which synthesizes the output image.

where IoU is the intersection-over-union score, and $i$ iterates over segments in $\mathbf{M}$ that have the same semantic class as $L_j$. The first term (mask IoU) measures the overlap of the segment shapes. The second term (context IoU) measures the similarity of the surrounding semantic layout. The use of semantic context helps retrieve compatible segments when surrounding context affects appearance.

### 4.3. Transformation

The transformation network $T$ is designed to transform the selected object segment $P_{\sigma(j)}$ to match $L_j$ via translation, rotation, scaling, and clipping. The transformation aims to align $P_{\sigma(j)}$ to $L_j$ while preserving the integrity of the object's appearance. $T(L, L_j^{mask}, P_{\sigma(j)}^{color})$ takes $L$, $L_j^{mask}$, and $P_{\sigma(j)}^{color}$ as input and produces a transformed image $\tilde{P}_{\sigma(j)}$ by applying a 2D affine transformation to $P_{\sigma(j)}^{color}$ [12]. We use a deep network rather than an analytical approach because a network can learn to preserve properties such as symmetry and upright orientation as needed, without hard-coding such properties as rules.

To train the network $T$, we need to generate segment pairs that will simulate the inconsistencies in shape, scale, and location that $T$ encounters at test time. For this reason, simply training $T$ to transform $P_i^{color}$ to match $P_i^{mask}$, for segments $P_i \in \mathbf{M}$, does not work: the requisite transformation is trivial. We therefore simulate misalignments by applying random affine transformations and cropping to $P_i^{color}$. Let $\hat{P}_i^{color}$ be the image produced by such transformation. The network $T$ is trained to align $\hat{P}_i^{color}$ with $P_i^{mask}$. The training loss for $T$ is

$$\mathcal{L}_T(\theta^T) = \sum_{P_i \in \mathbf{M}} \left\| P_i^{color} - T(P, P_i^{mask}, \hat{P}_i^{color}; \theta^T) \right\|_1.$$

This loss is defined over the color images rather than the mask because information in the color image is more specific and better constrains the transformation.

### 4.4. Canvas

After selecting and transforming object segments for the semantic layout $L$, we composite these segments onto a single canvas image. Let $\tilde{P}_{\sigma(j)}$ be the transformed segment for $L_j$. If all pairs $(\tilde{P}_{\sigma(i)}, \tilde{P}_{\sigma(j)})$ are disjoint, the canvas composition is trivial. If $\tilde{P}_{\sigma(i)}$ and $\tilde{P}_{\sigma(j)}$ overlap, we need to determine their order, since one of them will occlude the other. For example, when a building segment overlaps a sky segment, the sky segment should be occluded.

We train an ordering network to determine the front-back ordering of adjacent object segments. The architecture of the ordering network is based on VGG-19 [30]. Its output is a $c$-dimensional one-hot vector that indicates the semantic label of the segment that should be in front. When two segments overlap, we query the ordering network to determine their front-back order on the canvas $C$.

To train the network, we use the relative depth of adjacent semantic segments in the training set. This relative depth can be estimated from depth or stereo data provided with some datasets, such as Cityscapes and NYU [5, 29]. For datasets without such auxiliary information, such as ADE20K [39], we generate approximate depth maps using a network trained for this purpose [4]. The approximate depth maps are used to determine the relative depth order of adjacent segments in the training data, which are in turn used to train the ordering network. The ordering network is trained with a cross-entropy loss.

The boundaries of segments in the canvas are elided as described in Section 5.2.

## 5. Image Synthesis

The image synthesis network $f$ takes as input the semantic layout $L$, the canvas $C$, and a binary mask that indicates missing pixels in the canvas. The canvas $C$ provides raw material for synthesis, but is inadequate in itself: regions are typically missing, different segments are inconsistently illuminated and color-balanced, and a variety of boundary artifacts are apparent. Missing regions could be filled using an inpainting network [23, 35, 9], but this does not address other artifacts that are present in the canvas. We therefore design and train a dedicated network that takes both the canvas and the target semantic layout into account.

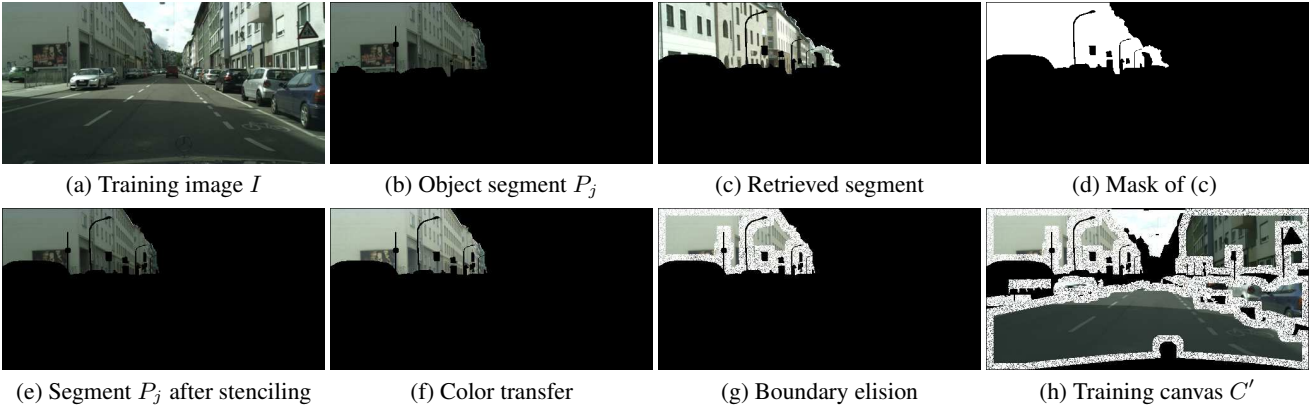

Figure 4. Generation of a simulated canvas during training. (a) Training image $I$. (b) Object segment $P_j$ extracted from the training image. (c) A corresponding segment retrieved from a different image in the training set. (d) Mask of the retrieved segment, used to stencil $P_j$. (e) The segment $P_j$ is stenciled with the mask of the retrieved segment. (f) Color transfer is applied to further modify the appearance of $P_j$. (g) Boundaries are elided to force the synthesis network to learn to synthesize content near boundaries. (h) A complete canvas $C'$ generated from image $I$ for training the synthesis network.

### 5.1. Network architecture

The architecture of the synthesis network $f$ is shown in Figure 3. The network has an encoder-decoder structure with skip connections. The encoder constructs a multi-scale representation of the input $(C, L)$. The decoder uses this representation to synthesize progressively finer feature maps, culminating in full-resolution output.

**Encoder.** Our encoder is based on VGG-19 [30]. The input is a tensor that collates $L$ and $C$. The network consists of five modules. Each module contains a number of convolutional layers [16] with layer normalization [1], ReLU [20], and average pooling. The first module has two convolutional layers, while each of the other modules have three. Each element in the encoder's output tensor has a receptive field of approximately $276 \times 276$. The encoder can thus capture long-range correlations that can help the decoder harmonize color, lighting, and texture.

**Decoder.** Our decoder is based on the cascaded refinement network (CRN) [2]. The network is a cascade of refinement modules. The input to each module is a concatenation of feature maps produced at the corresponding resolution by the encoder, feature maps produced by the preceding refinement module (if any), the canvas $C$ (appropriately resized), and the semantic layout $L$ (resized). Each refinement module contains two convolutional layers with layer normalization and Leaky ReLU [18].

### 5.2. Training

The image synthesis network $f$ is trained using simulated canvases that are generated to mimic artifacts that are encountered at test time. Given a semantic layout $L$ and a corresponding color image $I$ from the training set, we generate a simulated canvas $C'$ by applying stenciling, color transfer, and boundary elision to segments in $(I, L)$. The network $f$ is trained to take the pair $(C', L)$ and recover the original image $I$. Following [2], the network is trained using a perceptual loss based on feature activations in a pre-trained VGG-19 network [30]. The loss is

$$\mathcal{L}_f(\theta^f) = \sum_{(I,L)\in\mathcal{D}} \sum_l \lambda_l \|\Phi_l(I) - \Phi_l(f(C', L); \theta^f)\|_1,$$

where $\Phi_l$ is the feature tensor in layer $l$, and the weights $\{\lambda_l\}$ balance the terms. We use 'conv1_2', 'conv2_2', 'conv3_2', 'conv4_2', and 'conv5_2' layers in the loss.

We now review the generation of the simulated canvas $C'$, organized into a number of steps.

**Stenciling.** It is inevitable that the test-time canvas $C$ will contain missing regions. Thus the network $f$ must be trained on simulated canvases with realistic missing regions. We simulate missing regions by stenciling each segment in $(I, L)$ using a mask obtained from a different segment in the dataset. Specifically, for each segment $P_j$, we use the retrieval procedure described in Section 4.2 to retrieve a segment from a different image in the training set. The mask of that segment is then used to stencil $P_j$. This is illustrated in Figure 4 (b-e).

**Color transfer.** At test time, different segments composited onto the canvas will generally have inconsistent tone and illumination. To simulate these artifacts in the training canvas $C'$, we select 20% of the segments at random and apply color transfer [27]. Specifically, to modify the color distribution of a segment $P_j$ in $C'$, we randomly retrieve a segment $P_i$ with the same semantic class from $\mathbf{M}$ and transfer the color distribution from $P_i$ to $P_j$. This is illustrated in Figure 4(f).

|  | Cityscapes-coarse | Cityscapes-fine | Cityscapes→GTA5 | NYU | ADE20K | Mean |
| --- | --- | --- | --- | --- | --- | --- |
| SIMS > Pix2pix [11] | 94.2% | 98.1% | 95.7% | 94.9% | 87.6% | 94.1% |
| SIMS > CRN [2] | 93.9% | 74.1% | 84.5% | 89.1% | 88.9% | 86.1% |

Table 1. Results of blind randomized A/B tests. Each entry reports the percentage of comparisons in which an image synthesized by our approach (SIMS) was judged more realistic than a corresponding image synthesized by Pix2pix [11] or the CRN [2]. Chance is at 50%.

**Boundary elision.** The network $f$ should also be trained to naturally blend object boundaries. To encourage this, we randomly mask out 80% of pixels within a distance of $0.05h$ from a segment boundary. These are replaced by white pixels. The network is thus forced to learn to synthesize content near boundaries. This masking of interior boundary regions is illustrated in Figure 4(g).

Furthermore, inconsistencies along boundaries arise not only inside segments, but also outside. Consider a car composited onto a road. A typical salient artifact is the absence of shadow beneath the car. To encourage the network to learn to synthesize such shadows and other near-range inter-object effects, we also mask out pixels in $C'$ that lie outside an object segment within a distance of $0.05h$ from its boundary. These are replaced by black pixels. The network $f$ is forced to inpaint these exterior regions.

The same interior and exterior boundary elision steps are also applied at test time.

## 6. Experiments

**Datasets.** We conduct experiments on three semantic segmentation datasets: Cityscapes [5], NYU [29], and ADE20K [39]. The Cityscapes dataset contains images of urban street scenes. It provides 3K images with fine pixel-wise annotations and 20K images with coarse incomplete annotations for training. We train models separately for the fine and coarse regimes, and test on the 500 images in the validation set, which have both fine and coarse label maps. For the NYU dataset, we train on the first 1200 images and test on the remaining 249 images in the dataset. For ADE20K, we use outdoor images from the dataset; this yields 10K images for training and 1K images for testing.

**Perceptual experiments.** We adopt the experimental protocol of Chen and Koltun [2]. The protocol is based on large batches of blind randomized A/B tests deployed on the Amazon Mechanical Turk platform. We compare the presented approach to Pix2pix [11] and the CRN [2].

Table 1 reports the results. Each entry in the table reports the percentage of comparisons in which an image synthesized by our approach (SIMS) was judged more realistic than a corresponding image synthesized by Pix2pix or the CRN. Models trained on Cityscapes are tested in three conditions: 'Cityscapes-coarse' for models trained and tested on coarse input layouts, 'Cityscapes-fine' for models trained and tested on fine input layouts, and Cityscapes→GTA5 for models trained on fine Cityscapes layouts and then applied to semantic label maps from the GTA5 dataset [28]. (We use the 6K semantic layouts in the GTA5 validation set.) Note that chance is at 50%.

In all conditions, the presented approach outperforms the baselines. Across the five datasets, images synthesized by our approach were rated more realistic than images synthesized by Pix2pix and the CRN in 94% and 86% of comparisons, respectively.

We have also conducted time-limited pairwise comparisons, again following the protocol of Chen and Koltun [2]. The results are reported in Figure 6. Here each comparison pairs an image synthesized by one of the approaches versus the real reference image for the same semantic layout. In this case 50% is the equivalent of passing the visual Turing test. While none of the approaches achieves this, images synthesized by SIMS are more frequently mistaken for real ones. For example, after 1 second, the preference rate for SIMS>Real in the Cityscapes-coarse condition is 25.2%, versus 4.0% for CRN>Real and 3.8% for Pix2pix>Real. After 1 second in the Cityscapes-fine condition, the preference rate for SIMS>Real is 27.8%, versus 15.2% for CRN>Real and 1.9% for Pix2pix>Real.

**Semantic segmentation accuracy.** Next we analyze the realism of synthesized images using a different protocol. Given a semantic layout $L$, we use one of the evaluated approaches to synthesize an image $I$. This image is then given as input to a pretrained semantic segmentation network. (We use the PSPNet [38].) This network produces a semantic layout $\hat{L}$ as output. We then compare $\hat{L}$ to the original layout $L$. In principle, the closer these are, the more realistic the intermediate synthesized image $I$ can be assumed to be [11]. We evaluate the similarity of $L$ and $\hat{L}$ using two measures: intersection over union (IoU) and overall pixel accuracy. These measures are averaged over all images in the test set of each dataset.

The results are reported in Table 2. Images synthesized by our approach can be more accurately parsed by the PSPNet than images synthesized by Pix2pix or the CRN. The differences on Cityscapes-coarse and ADE20K are dra-matic. Note that this experimental procedure also evaluates conformance with the input semantic layout: if the image synthesized by an approach is realistic but does not conform to the input layout, it will be penalized by this protocol.

|  | Cityscapes-coarse | | Cityscapes-fine | | ADE20K | |
| --- | --- | --- | --- | --- | --- | --- |
|  | IoU | Accuracy | IoU | Accuracy | IoU | Accuracy |
| Reference | 84.0% | 90.1% | 71.7 % | 81.6% | 37.9% | 48.5% |
| Pix2pix [11] | 30.1% | 34.2% | 31.0% | 37.9% | 16.1% | 26.6% |
| CRN [2] | 28.5% | 36.2% | **51.3%** | 61.4% | 23.1% | 30.7% |
| SIMS | **56.3%** | **65.6%** | 51.4% | **65.5%** | **38.4%** | **50.1%** |

Table 2. Images synthesized by different approaches are given to a pretrained semantic segmentation network (PSPNet [38]). Its output is compared to the semantic layout that was used as input for image synthesis. IoU and pixel accuracy are averaged across each dataset. Higher is better. 'Reference' is the value achieved by using the real images as input to the PSPNet.

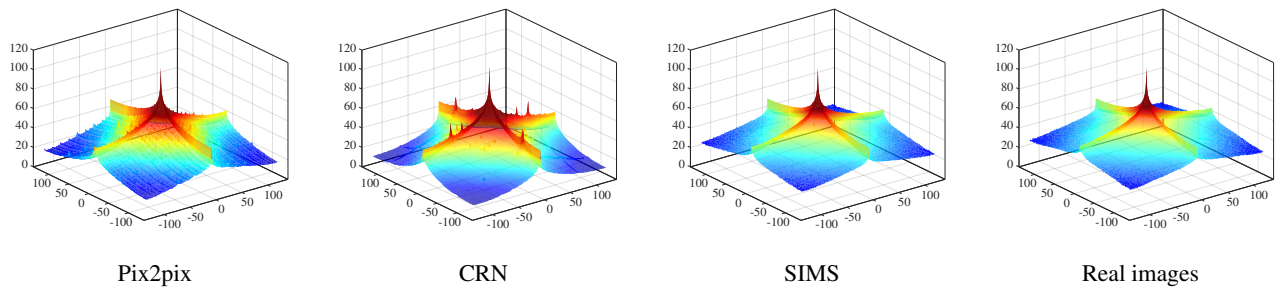

| Pix2pix | CRN | SIMS | Real images |

Figure 5. Mean power spectra over the ADE20K dataset. Magnitude is on a logarithmic scale. We compare the mean power spectra of images synthesized by Pix2pix, CRN, and SIMS to the mean power spectrum of real images from the ADE20K test set. The mean power spectrum of images synthesized by SIMS is virtually indistinguishable from the mean power spectrum of real images, while the mean power spectra of images synthesized by Pix2pix and the CRN are characterized by spurious spikes. Zoom in for details.

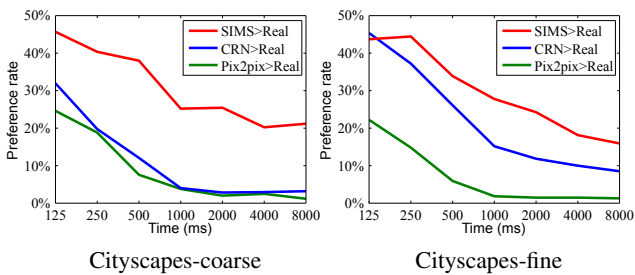

Figure 6. Time-limited pairwise comparisons versus real images. 50% is the equivalent of passing the visual Turing test.

**Image statistics.** We now analyze the realism of synthesized images in terms of low-level image statistics. We consider the mean power spectrum of synthesized images across a given dataset, versus corresponding real images from the dataset [36]. Figure 5 shows the mean power spectra of images synthesized by Pix2pix, CRN, and SIMS, averaged across the ADE20K dataset. The mean power spectrum of real ADE20K images is shown for reference. As can be seen in the figure, the mean power spectrum of images synthesized by our approach is virtually indistinguishable from the mean power spectrum of real images. In contrast, the mean power spectra of images synthesized by

Pix2pix and CRN are clearly spiky, with many spurious local maxima that are not present in real images.

**Qualitative results.** Figure 7 shows a number of images synthesized by Pix2pix, CRN, and SIMS, trained on the Cityscapes dataset. Results are shown in the three conditions summarized earlier: Cityscapes-coarse, Cityscapes-fine, and Cityscapes→GTA5. Figure 8 shows examples of synthesized images for the NYU and ADE20K datasets. Additional results are provided in the supplement.

**Diversity.** The presented approach can be easily extended to synthesize a diverse collection of images. To this end, the retrieval stage described in Section 4.2 can be modified to retrieve not a single segment that maximizes the presented score, but a random segment among the top $k$ segments that maximize the score across the dataset. The retrieved segment for each semantic region $L_j$ can be randomized in this fashion. Given an input layout, the synthesis process can be repeated to synthesize as many corresponding images as desired. Results of this process are shown in the supplement.

## 7. Conclusion

We have presented a semi-parametric approach to photographic image synthesis from semantic layouts. Experiments demonstrate that the presented approach (SIMS) produces considerably more realistic images than recent purely parametric techniques. Note that the quality of SIMS is in a sense lower-bounded by the performance of parametric methods: if the memory bank is not useful, the network $f$ can simply ignore the canvas and perform parametric synthesis based on the input semantic layout.

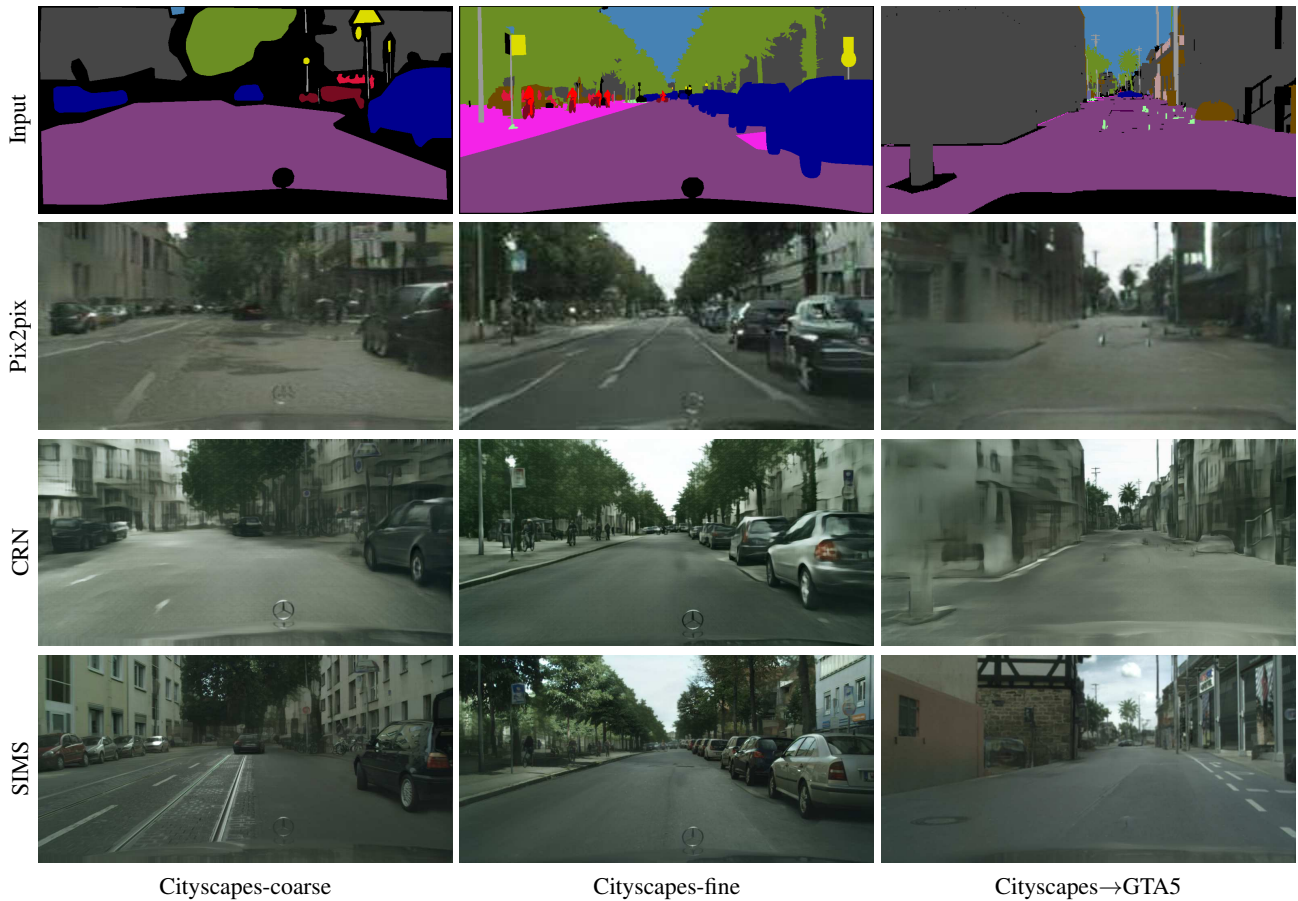

Figure 7. Images synthesized by Pix2pix, CRN, and SIMS. This figure shows results produced by models trained on the Cityscapes dataset.

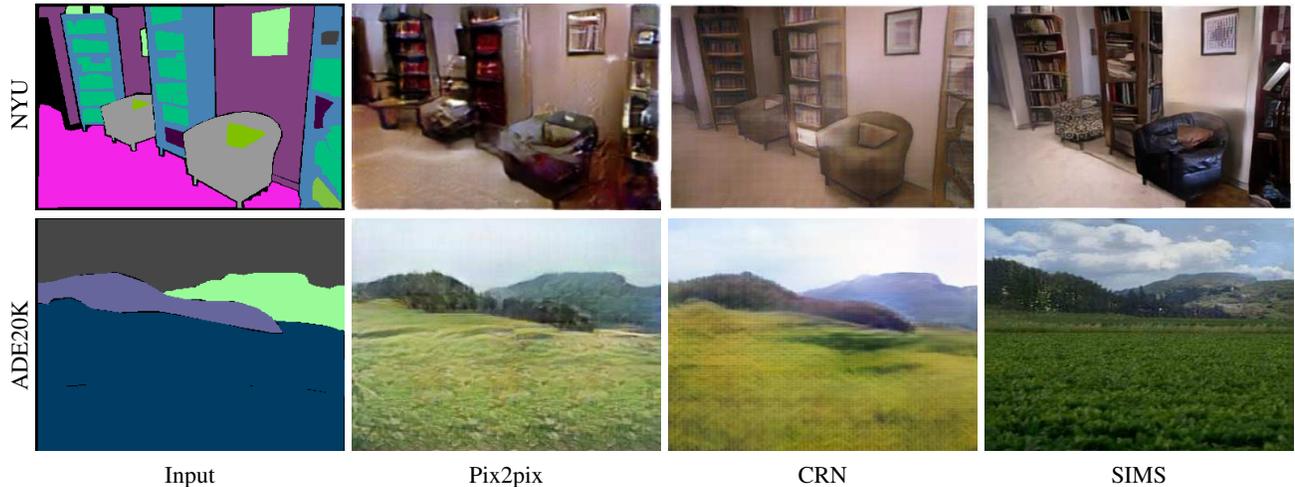

Figure 8. Images synthesized by models trained on the NYU and ADE20K datasets.

Many interesting problems are left open for future work. First, our implementation is significantly slower than purely parametric methods; more efficient data structures and algorithms should be explored. Second, other forms of input can be used, such as semantic instance segmentation or textual descriptions. Third, the presented pipeline is not trained end-to-end. Lastly, applying semi-parametric techniques to video synthesis is an exciting frontier.